\documentclass{article}
\usepackage[nonatbib,final]{neurips_2019}

\usepackage[utf8]{inputenc} 
\usepackage[T1]{fontenc}    
\usepackage{hyperref}       
\usepackage{url}            
\usepackage{booktabs}       
\usepackage{amsmath}
\usepackage{amsfonts}       
\usepackage{nicefrac}       
\usepackage{microtype}      
\usepackage{graphicx}

\DeclareMathOperator*{\argmax}{arg\,max}

\title{Well-calibrated Model Uncertainty with Temperature Scaling for Dropout Variational Inference}

\author{%
  Max-Heinrich~Laves \\
  \And
  Sontje~Ihler \\
  \And
  Karl-Philipp~Kortmann \\
  \And
  Tobias~Ortmaier \\
  \AND \vspace*{-1em}
  \\ 
  Institute of Mechatronic Systems\\
  Leibniz University Hannover, Germany\\
  \texttt{\{laves,ihler,kortmann,ortmaier\}@imes.uni-hannover.de}
}

\begin{document}

\maketitle
\begin{abstract}
Model uncertainty obtained by variational Bayesian inference with Monte Carlo dropout is prone to miscalibration.
The uncertainty does not represent the model error well.
In this paper, temperature scaling is extended to dropout variational inference to calibrate model uncertainty.
Expected uncertainty calibration error (UCE) is presented as a metric to measure miscalibration of uncertainty.
The effectiveness of this approach is evaluated on CIFAR-10/100 for recent CNN architectures.
Experimental results show, that temperature scaling considerably reduces miscalibration by means of UCE and enables robust rejection of uncertain predictions.
The proposed approach can easily be derived from frequentist temperature scaling and yields well-calibrated model uncertainty.
It is simple to implement and does not affect the model accuracy.
\end{abstract}
\section{Introduction}

For safety-critical vision tasks such as autonomous driving or computer-aided diagnosis, it is essential to consider the prediction \emph{uncertainty} of deep learning models.
Bayesian neural networks and recent advances in their approximation provide the mathematical tools for quantification of uncertainty \cite{Bishop2006,Kingma2013}.
One practical approximation is variational inference with Monte Carlo (MC) dropout \cite{Gal2016}.
It is applied to obtain epistemic uncertainty, which is caused by uncertainty in the model weights due to training with data sets of limited size \cite{Bishop2006,Kendall2017}.
However, it tends to be miscalibrated, i.\,e. the uncertainty does not correspond well to the model error\,\cite{Guo2017,Gal2017}.

First, we consider the problem of miscalibration of the frequentist approach:
The weights of a deep model are obtained by maximum likelihood estimation \cite{Bishop2006}, and the normalized output likelihood for an unseen test input does not consider uncertainty in the weights \cite{Kendall2017}.
The likelihood is generally unjustifiably high \cite{Guo2017}, and can be misinterpreted as high prediction \emph{confidence}.
This miscalibration can also be observed for model uncertainty provided by MC dropout variational inference.
However, calibrated uncertainty is essential as miscalibration can lead to decisions with fatal consequences in the aforementioned task domains.

Overconfident predictions of neural networks have been addressed by entropy regularization techniques.
Szegedy et al.\ present label smoothing as regularization of models during supervised training for classification \cite{Szegedy2016}.
They state that a model trained with one-hot encoded labels is prone to becoming overconfident about its predictions, which causes overfitting and poor generalization.
Pereyra et al.\ link label smoothing to \emph{confidence penalty} (CP) and propose a simple way to prevent overconfident networks \cite{Pereyra2017}.
Low entropy output distributions are penalized by adding the negative entropy to the training objective.
However, the referred works do not apply entropy regularization to the calibration of confidence or uncertainty.
In the last decades, several non-parametric and parametric calibration approaches such as isotonic regression \cite{Zadrozny2002} or Platt scaling \cite{Platt1999} have been presented.
Recently, \emph{temperature scaling} (TS) has been demonstrated to lead to well-calibrated model likelihood in non-Bayesian deep neural networks \cite{Guo2017}.
It uses a single scalar to smooth the softmax output and regularize the entropy.
Scaling has also been introduced to approximate categorical distributions by the Gumbel-Softmax or Concrete distribution \cite{Jang2016,Maddison2016}.

Our work extends temperature scaling to variational Bayesian inference with dropout to obtain well-calibrated model uncertainty.
The main contributions of this paper are 1. definition for perfect calibration of uncertainty and definition for the expected uncertainty calibration error, 2. the derivation of temperature scaling for dropout variational inference, and 3. experimental results of different network architectures on CIFAR-10/100 \cite{Krizhevsky2009}, that demonstrate the improvement of calibration by the proposed method and superiority over confidence penalty.
By using temperature scaling together with Bayesian inference, we expect better calibrated uncertainty.
To the best of our knowledge, temperature scaling has not yet been used to calibrate model uncertainty in variational Bayesian inference.
Our code is available at: \href{https://github.com/mlaves/bayesian-temperature-scaling}{\texttt{https://github.com/mlaves/bayesian-temperature-scaling}}.

\section{Methods}

The presented approach for obtaining well-calibrated uncertainty is applied to a general multi-class classification task.
Let input $ \mathbf{x} \in \mathcal{X} $ be a random variable with corresponding label $ y \in \mathcal{Y} = \{1, \ldots , C\} $.
Let $ \mathbf{f}_{\mathbf{w}}(\mathbf{x}) $ be the output (logits) of a neural network with weight matrices $ \mathbf{w} $, and with model likelihood $ p( y \! = \! c \,\vert\,  \mathbf{f}_{\mathbf{w}}(\mathbf{x}) ) $ for class $ c $, which is sampled from a probability vector $ \mathbf{p} = \boldsymbol{\sigma}_{\mathrm{SM}}(\mathbf{f}_{\mathbf{w}}(\mathbf{x})) $, obtained by passing the model output through the softmax function $ \boldsymbol{\sigma}_{\mathrm{SM}}(\cdot) $.
From a frequentist perspective, the softmax likelihood is often interpreted as confidence of prediction.
Throughout this paper, we follow this definition.
However, due to optimizing the weights $ \mathbf{w} $ via minimization of the negative log-likelihood of $ p( y \,\vert\, \mathbf{f}_{\mathbf{w}}(\mathbf{x}) ) $, modern deep models are prone to overly confident predictions and are therefore miscalibrated \cite{Guo2017,Gal2017}.

Let $ \hat{y} = \argmax \mathbf{p} $ be the most likely class prediction of input $ \mathbf{x} $ with likelihood $ \hat{p} = \max \mathbf{p} $ and true label $ y $. Then, following Guo et al. \cite{Guo2017}, \emph{perfect calibration} is defined as
\begin{equation}
    \mathbb{P} \left( \hat{y} = y \,\vert\, \hat{p} = q \right) = q , \quad \forall q \in \left[ 0, 1 \right] .
    \label{eq:perfect_calibration}
\end{equation}
To determine model uncertainty, dropout variational inference is performed by training the model $ \mathbf{f}_{\mathbf{w}} $ with dropout \cite{Srivastava2014} and using dropout at test time to sample from the approximate posterior by performing $ N $ stochastic forward passes~\cite{Gal2016,Kendall2017}.
This is also referred to as MC dropout.
In MC dropout, the final probability vector is obtained by MC integration:
\begin{equation}
    \mathbf{p} (\mathbf{x}) = \frac{1}{N} \sum_{i=1}^{N} \boldsymbol{\sigma}_{\mathrm{SM}} \left( \mathbf{f}_{\mathbf{w}_{i}} (\mathbf{x}) \right) .
\end{equation}
Entropy of the softmax likelihood is used to describe \emph{uncertainty} of prediction \cite{Kendall2017}.
In contrast to confidence as a measure of goodness of prediction, uncertainty takes into account the likelihoods of all $ C $ classes.
We introduce normalization to scale the values to a range between $ 0 $ and $ 1 $:
\begin{equation}
    \tilde{\mathcal{H}}(\mathbf{p}) := - \frac{1}{\log C} \sum_{c=1}^{C} p^{(c)} \log p^{(c)} ~ , \quad \tilde{\mathcal{H}} \in \left[0, 1\right] .
    \label{eq:norm_entropy}
\end{equation}
From Eq.\,(\ref{eq:perfect_calibration}) and Eq.\,(\ref{eq:norm_entropy}), we define \emph{perfect calibration of uncertainty} as
\begin{equation}
    \mathbb{P} ( \hat{y} \neq y \,\vert\, \tilde{\mathcal{H}}( \mathbf{p} ) = q ) = q , \quad \forall q \in \left[0, 1\right] .
\end{equation}
That is, in a batch of inputs all predicted with uncertainty of e.g. $ 0.2 $, a top-1 error of $ 20\,\% $ is expected.

\subsection{Expected Uncertainty Calibration Error (UCE)}

A popular way to quantify miscalibration of neural networks with a scalar value is the expectation of the difference between predicted softmax likelihood $ \hat{p} $ and accuracy
\begin{equation}
    \mathbb{E}_{\hat{p}}\left[ \, \left| \mathbb{P} \left( \hat{y} = y \,\vert\, \hat{p} = q \right) - q \right| \, \right], \quad \forall q \in \left[ 0, 1 \right] ,
    \label{eq:ece}
\end{equation}
which can be approximated by the Expected Calibration Error (ECE) \cite{Naeini2015,Guo2017}.
The output of a neural network is partitioned into $ M $ bins with equal width and a weighted average of the difference between accuracy and confidence (softmax likelihood) is taken:
\begin{equation}
    \mathrm{ECE} = \sum_{m=1}^{M} \frac{\left| B_{m} \right|}{n} \, \big| \mathrm{acc}(B_{m}) - \mathrm{conf}(B_{m}) \big| ~ ,
\end{equation}
with total number of inputs $ n $ and set of indices $ B_{m} $ of inputs whose confidence falls into that bin (see \cite{Guo2017} for more details).
We propose the following slightly modified notion of Eq.\,(\ref{eq:ece}) to quantify miscalibration of uncertainty:
\begin{equation}
    \mathbb{E}_{\tilde{\mathcal{H}}} [ \, \vert \mathbb{P} ( \hat{y} \neq y \,\vert\, \tilde{\mathcal{H}}( \mathbf{p} ) = q ) - q \vert  \, ], \quad \forall q \in \left[ 0, 1 \right] .
\end{equation}
We refer to this as Expected Uncertainty Calibration Error (UCE) and approximate analogously with
\begin{equation}
    \mathrm{UCE} := \sum_{m=1}^{M} \frac{\left| B_{m} \right|}{n} \big| \mathrm{err}(B_{m}) - \mathrm{uncert}(B_{m}) \big| ~ .
    \label{eq:uce}
\end{equation}
See §\,\ref{app:uce} for definitions of $ \mathrm{err}(B_{m}) $ and $ \mathrm{uncert}(B_{m}) $.

\subsection{Temperature Scaling for Dropout Variational Inference}

\begin{figure}
    \centering
    \includegraphics[width=0.95\textwidth]{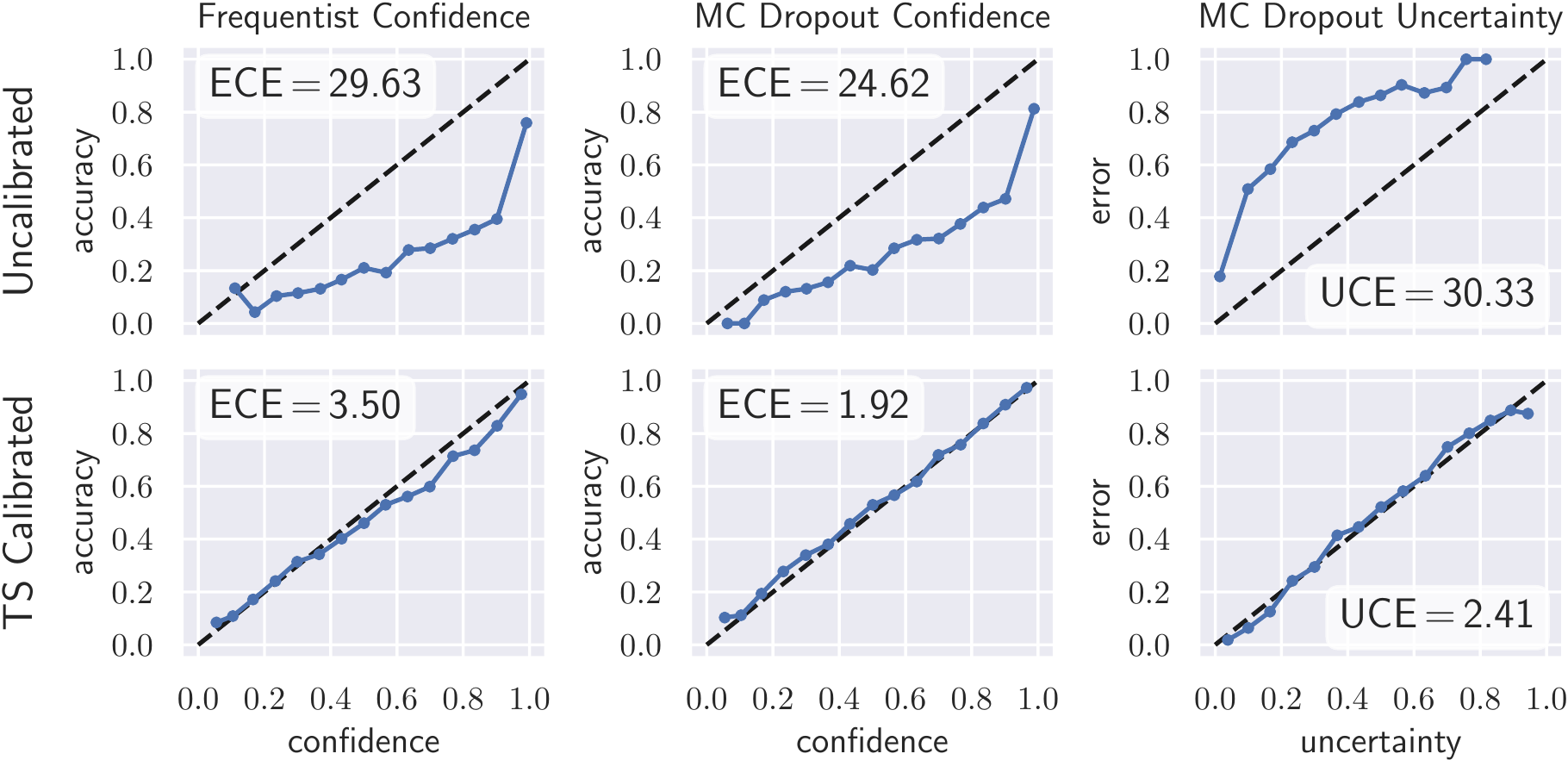}
    \caption{Reliability diagrams ($ M = 15 $ bins) for ResNet-101 on CIFAR-100. Top row: Uncalibrated frequentist confidence (left), and confidence and uncertainty obtained by dropout variational inference (right). Bottom row: Results from calibration with TS. Dashed lines denote perfect calibration.}
    \label{fig:reliability}
\end{figure}

State-of-the-art deep neural networks are generally miscalibrated with regard to softmax likelihood \cite{Guo2017}.
However, when obtaining model uncertainty with dropout variational inference, this also tends to be not well-calibrated \cite{Gal2017}.
Fig.\,\ref{fig:reliability} (top row) shows reliability diagrams \cite{Niculescu2005} for uncalibrated ResNet-101 \cite{He2016} trained on CIFAR-100 \cite{Krizhevsky2009}.
The divergence from the identity function reveals miscalibration.

In this work, dropout is inserted before the last layer with fixed dropout probability of $ 0.5 $ as in \cite{Gal2016}.
Temperature scaling with $ T > 0 $ is inserted before final softmax activation and before MC integration:
\begin{equation}
    \hat{\mathbf{p}} (\mathbf{x}) = \frac{1}{N} \sum_{i=1}^{N} \boldsymbol{\sigma}_{\mathrm{SM}} \left( T^{-1} \mathbf{f}_{\mathbf{w}_{i}} (\mathbf{x}) \right) .
\end{equation}
$ T $ is optimized with respect to negative log-likelihood while performing MC dropout on the validation set.
This is equivalent to maximizing the entropy of $ \hat{\mathbf{p}} $ \cite{Guo2017}. See §\,\ref{app:ts} for more details on $ T $.

\section{Experiments \& Results}
\label{sec:results}

The experimental results of the proposed method are presented threefold:
First, TS is used to calibrate confidence and uncertainty obtained by MC dropout;
second, TS calibration is compared with calibration by entropy regularization using confidence penalty; and finally, uncertain predictions are rejected based on well-calibrated uncertainty.
Details on the training procedure can be found in §\,\ref{app:training}.

\subsection{Uncertainty Calibration}

Tab.\,\ref{tab:results} reports test set results for different networks \cite{He2016,Huang2017} and data sets used to evaluate the performance of temperature scaling for dropout variational inference.
The proposed UCE metric is used to quantify calibration of uncertainty.
Fig.\,\ref{fig:reliability} shows reliability diagrams \cite{Niculescu2005} for different calibration scenarios of ResNet-101 \cite{He2016} on CIFAR-100.
For MC dropout $ N=25 $ forward passes are performed.
Uncalibrated ECE shows, that MC dropout already reduces miscalibration of model likelihood by up to $ 6.6 $ percentage points. With TS calibration, MC dropout reduces ECE by 45--66\,\% and UCE drops drastically (especially for  larger networks). This illustrates the magnitude of how much TS calibration benefits from Bayesian inference using MC dropout. Additional reliability diagrams showing similar results can be found in §\,\ref{app:rel_diag}.
\begin{table}
    \small
    \centering
    \caption{ECE and UCE test set results in \% ($ M=15 $ bins). 0\,\% means perfect calibration. In TS calibration with MC dropout the same value of $ T $ was used to report both ECE and UCE.}
    \begin{tabular}{cccccccc}
        \toprule
         & & \multicolumn{3}{c}{Uncalibrated} & \multicolumn{3}{c}{TS Calibrated} \\
        \cmidrule(lr){3-5} \cmidrule(lr){6-8}
         & & Freq. & \multicolumn{2}{c}{MC Dropout} & Freq. & \multicolumn{2}{c}{MC Dropout} \\
        \cmidrule(lr){3-3} \cmidrule(lr){4-5} \cmidrule(lr){6-6} \cmidrule(lr){7-8}
        Data Set & Model & ECE & ECE & UCE & ECE & ECE & UCE \\
        \midrule
        CIFAR-10 & ResNet-18 & 8.95 & 8.41 & 7.60 & 1.40 & \textbf{0.47} & \textbf{5.27} \\
        CIFAR-100 & ResNet-101 & 29.63 & 24.62 & 30.33 & 3.50 & \textbf{1.92} & \textbf{2.41} \\
        CIFAR-100 & DenseNet-169 & 30.62 & 23.98 & 29.62 & 6.10 & \textbf{2.89} & \textbf{2.69} \\
        \bottomrule \\
    \end{tabular}
    \label{tab:results}
\end{table}

\subsection{Temperature Scaling vs. Confidence Penalty}
\label{sec:conf_penalty}

\begin{figure}
    \centering
    \includegraphics[width=0.9\textwidth]{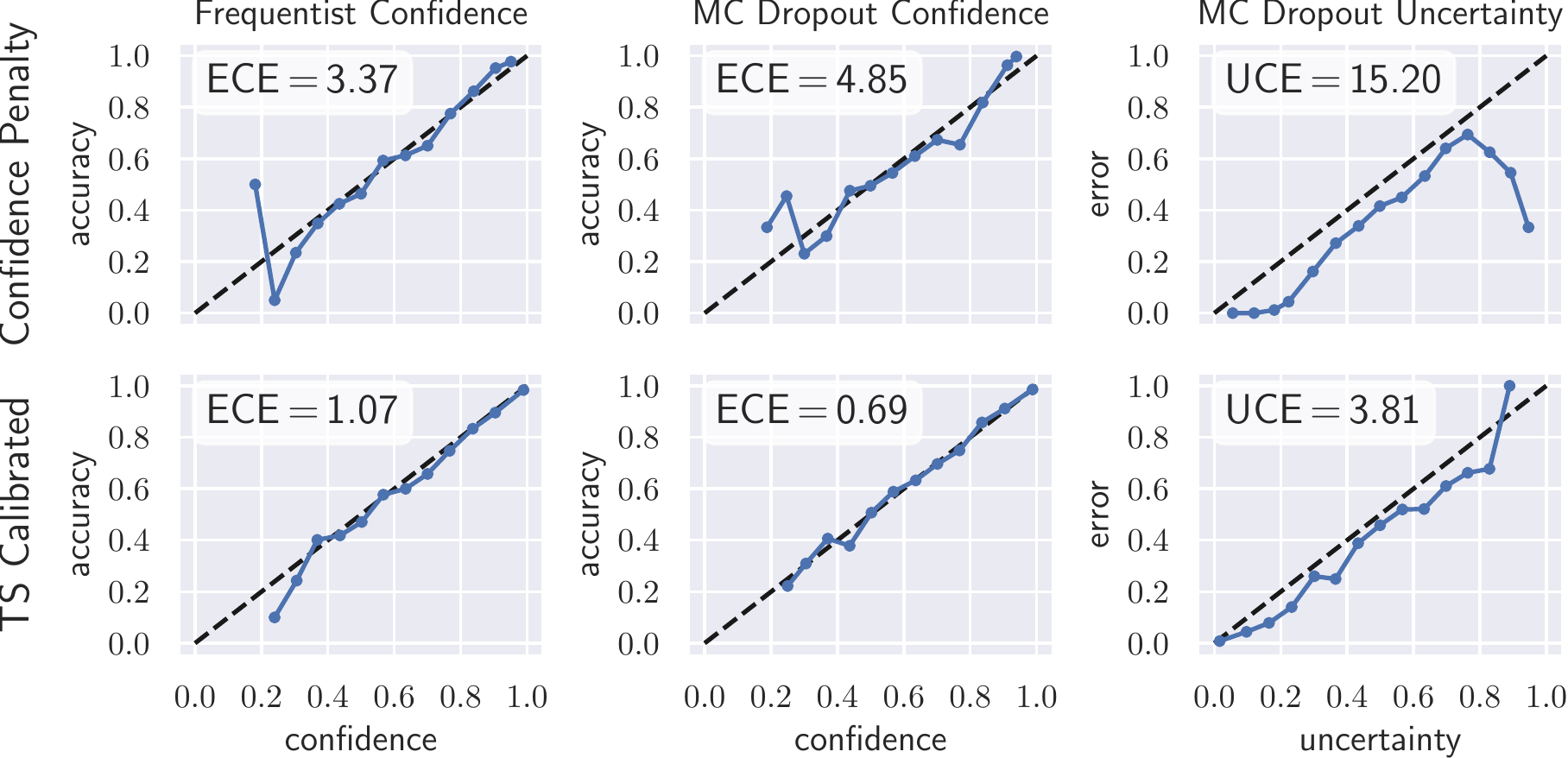}
    \caption{Reliability diagrams ($ M = 15 $ bins) for DenseNet-121 on CIFAR-10. Top row: Training with confidence penalty. Bottom row: TS calibrated (trained without confidence penalty).}
    \label{fig:conf_penalty}
\end{figure}
Low entropy output distributions are penalized by adding the negative entropy $ \mathcal{H} $ of the softmax output to the negative log-likelihood training objective, weighted by an additional hyperparameter $ \beta $.
This leads to the following optimization function:
\begin{equation}
    \mathcal{L}_{\mathrm{CP}}(\mathbf{w}) = - \sum_{\mathcal{X}, \mathcal{Y}} \log \mathbf{p}_{\mathbf{w}} (\mathbf{y} \vert \mathbf{x}) - \beta \, \mathcal{H} \left( \mathbf{p}_{\mathbf{w}}(\mathbf{y} \vert \mathbf{x}) \right) ~ .
    \label{eq:conf_penalty}
\end{equation}
We reproduce the experiment of Pereyra et al.\ on supervised image classification \cite{Pereyra2017} and compare the goodness of calibration of confidence and uncertainty to our presented approach.
DenseNet-121 with dropout is trained on CIFAR-10 as described in §\,\ref{app:training}.
We fix $ \beta = 0.1 $ for CP loss and omit data augmentation for this experiment as presented in \cite{Pereyra2017}.

Fig.\,\ref{fig:conf_penalty} compares training with confidence penalty to our approach.
CP reduces miscalibration ($ \mathrm{ECE} = 5.20\,\% $ without CP vs. $ \mathrm{ECE} = 3.37\,\% $ with CP for DenseNet-121).
However, it is not as effective as TS and still produces largely miscalibrated uncertainty.
A combination of CP during training and subsequent TS is conceivable and could possibly lead to an even better calibration.
We have not followed this approach yet.

\subsection{Example: Rejection of Uncertain Predictions}

\begin{figure}
    \centering
    \includegraphics[width=0.95\textwidth]{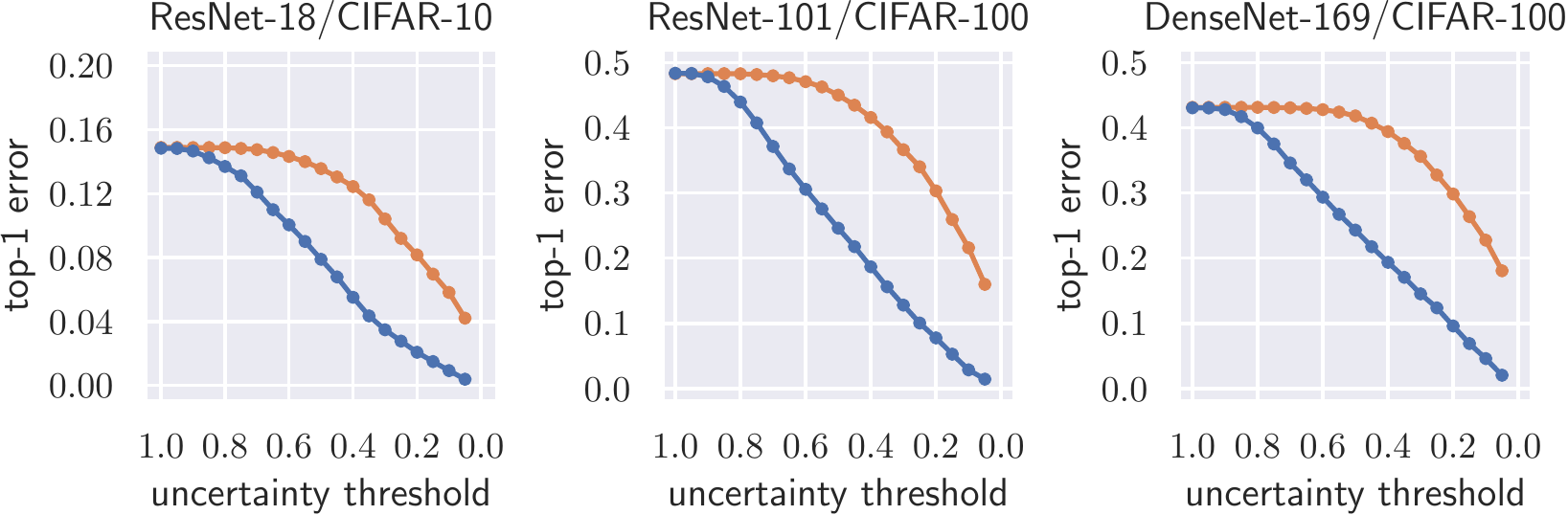}
    \caption{The effect of the uncertainty threshold $ \mathcal{H}_{\mathrm{max}} $ on the test set error for the rejection of uncertain predictions (orange: uncalibrated, blue: TS calibrated). As $ \mathcal{H}_{\mathrm{max}} $  decreases, more uncertain predictions are rejected, which results in a lower error.}
    \label{fig:uncert_thresh}
\end{figure}
An example application of well-calibrated prediction uncertainty is the rejection of uncertain predictions.
We define an uncertainty threshold $ \mathcal{H}_{\mathrm{max}} $ and reject all predictions from the test set where $ \tilde{\mathcal{H}}(\mathbf{p}) > \mathcal{H}_{\mathrm{max}} $.
A decrease in false predictions of the remaining test set is expected.
Fig.~\ref{fig:uncert_thresh} shows the top-1 error as a function of decreasing $ \mathcal{H}_{\mathrm{max}} $.
For both uncalibrated and calibrated uncertainty, decreasing $ \mathcal{H}_{\mathrm{max}} $ reduces the top-1 error.
Using calibrated uncertainty, the relationship is almost linear (for $ \mathcal{H}_{\mathrm{max}} < 0.8 $), allowing robust rejection of uncertain predictions.

\section{Conclusion}

Temperature scaling calibrates uncertainty obtained by dropout variational inference with high effectiveness.
The experimental results confirm the hypothesis that the presented approach yields better calibrated uncertainty.
In addition, substantially better calibrated softmax probability was achieved.
MC dropout TS is simple to implement, more effective than confidence penalty during training and the scaling does not change the maximum of the output of a network, thus model accuracy is not compromised.
Therefore, it is an obvious choice in Bayesian deep learning with dropout variational inference because well-calibrated uncertainties are of utmost importance for safety-critical decision-making.
However, there are many factors (e.\,g. network architecture, weight decay, dropout configuration) influencing the uncertainty in Bayesian deep learning that have not been discussed in this paper and are open to future work.

\subsubsection*{Acknowledgments}

This work has received funding from European Union EFRE projects \emph{OPhonLas} and \emph{ProMoPro}.
We thank the reviewers for their helpful comments.

\small

\bibliography{literature}
\bibliographystyle{unsrt}

\newpage
\normalsize
\appendix
\section{Appendix}
\label{appendix}

\subsection{Expected Uncertainty Calibration Error}
\label{app:uce}

We restate the definition of Expected Uncertainty Calibration Error (UCE) from Eq.\,(\ref{eq:uce}):
\begin{equation*}
    \mathrm{UCE} = \sum_{m=1}^{M} \frac{\left| B_{m} \right|}{n} \big| \mathrm{err}(B_{m}) - \mathrm{uncert}(B_{m}) \big| ~ .
\end{equation*}
The error per bin is defined as
\begin{equation}
    \mathrm{err}(B_{m}) = \frac{1}{\left| B_{m} \right| } \sum_{i \in B_{m}} \mathbf{1} (\hat{y}_{i} \neq y) ~ ,
\end{equation}
where $ \mathbf{1} (\hat{y}_{i} \neq y) = 1 $ and $ \mathbf{1} (\hat{y}_{i} = y) = 0 $.
Uncertainty per bin is defined as
\begin{equation}
    \mathrm{uncert}(B_{m}) = \frac{1}{\left| B_{m} \right| } \sum_{i \in B_{m}} \tilde{\mathcal{H}} (\mathbf{p}_{i}) ~ .
\end{equation}

\subsection{Temperature Scaling with Monte Carlo Dropout}
\label{app:ts}

Temperature scaling with MC dropout variational inference is derived by closely following the derivation of frequentist temperature scaling in the appendix of \cite{Guo2017}.
Let $ \left\{ \mathbf{z}_{1,j}, \ldots , \mathbf{z}_{N,j} \right\} $ be a set of logit vectors obtained by MC dropout with $ N $ stochastic forward passes for each input $ \mathbf{x}_{j} \in \left\{ \mathbf{x}_{1}, \ldots , \mathbf{x}_{M} \right\} $
with true labels $ \left\{ y_{1}, \ldots , y_{M} \right\} $.
Temperature scaling is the solution $ \hat{p} $ to entropy maximization
\begin{equation}
    \underset{\hat{p}}{\max} ~ - \frac{1}{N} \sum_{i=1}^{N} \sum_{j=1}^{M} \sum_{c=1}^{C} \hat{p} \left( \mathbf{z}_{i,j} \right)^{(c)} \log \hat{p} \left( \mathbf{z}_{i,j} \right)^{(c)} ,
\end{equation}
subject to
\begin{equation}
    \hat{p} (\mathbf{z}_{i,j})^{(c)} \geq 0 \quad \forall i,j,c ~ ,
\end{equation}
\begin{equation}
    \sum_{c=1}^{C} \hat{p} (\mathbf{z}_{j})^{(c)} = 1 \quad \forall j ~ ,
\end{equation}
\begin{equation}
    \frac{1}{N} \sum_{i=1}^{N} \sum_{j=1}^{M} z_{i,j}^{(y_{j})} = \frac{1}{N} \sum_{i=1}^{N} \sum_{j=1}^{M} \sum_{c=1}^{C} z_{i,j}^{(c)} \hat{p} ( \mathbf{z}_{i,j})^{(c)} .
\end{equation}
Guo et al. solve this constrained optimization problem with the method of Lagrange multipliers.
We skip reviewing their proof as one can see that the solution to $ \hat{p} $ in the case of MC dropout integration provides
\begin{align}
    \frac{1}{N} \sum_{i=1}^{N} \hat{p}_{i} \left( \mathbf{z}_{j} \right)^{(c)} &= \frac{1}{N} \sum_{i=1}^{N} \frac{e^{\lambda z_{i,j}^{(c)}}}{\sum_{\ell=1}^{C} e^{\lambda z_{i,j}^{(\ell)} }} \\
    &= \frac{1}{N} \sum_{i=1}^{N} \boldsymbol{\sigma}_{\mathrm{SM}} \left( \lambda \mathbf{f}_{\mathbf{w}_{i}} ( \mathbf{x}_{j} )\right)^{(c)} ,
\end{align}
which recovers temperature scaling for $ \lambda = T^{-1} $ \cite{Guo2017}.
$ T $ is optimized with respect to negative log-likelihood on the validation set using MC dropout.

\subsection{Training Settings}
\label{app:training}

The model implementations from PyTorch 1.2 \cite{Paszke2017} are used and trained with following settings:
\begin{itemize}
    \item batch size of $ 256 $
    \item AdamW optimizer \cite{Loshchilov2019} with initial learn rate of $ 0.01 $ and $ \beta_{1} = 0.9, \beta_{2} = 0.999 $
    \item weight decay of $ 0.01 $
    \item negative-log likelihood (cross entropy) loss
    \item reduce-on-plateau learn rate scheduler (patience of 10 epochs) with factor of $ 0.1 $
    \item additional validation set is randomly extracted from the training set (5000 samples)
    \item dropout with probability of $ 0.5 $ before the last linear layer was used in all models during training
    \item in MC dropout, $ N = 25 $ forward passes with dropout probability of $ 0.5 $ were performed
\end{itemize}
Code is available at: \href{https://github.com/mlaves/bayesian-temperature-scaling}{\texttt{https://github.com/mlaves/bayesian-temperature-scaling}}.

\subsection{Additional Reliability Diagrams}
\label{app:rel_diag}

In this section, reliability diagrams for the other data set/model combinations from Tab.\,\ref{tab:results} are visualized to provide additional insight into the calibration performance.
The proposed method is able to calibrate all models with respect to both UCE and ECE across all bins.
\begin{figure}[h]
    \centering
    \includegraphics[width=0.9\textwidth]{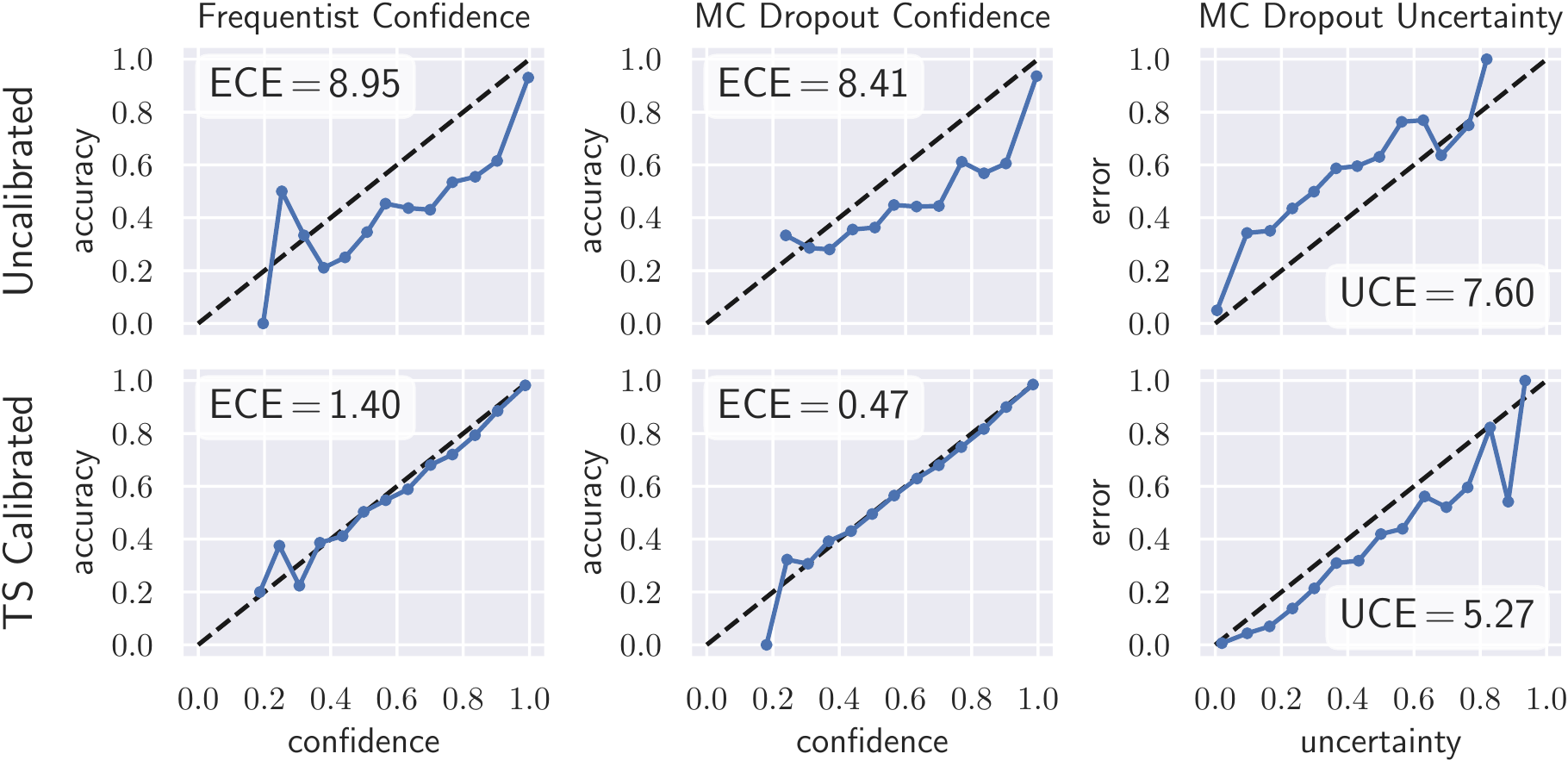}
    \caption{Reliability diagrams ($ M = 15 $ bins) for ResNet-18 on CIFAR-10.}
    \label{fig:reliability_resnet18}
\end{figure}

\begin{figure}[h]
    \centering
    \includegraphics[width=0.9\textwidth]{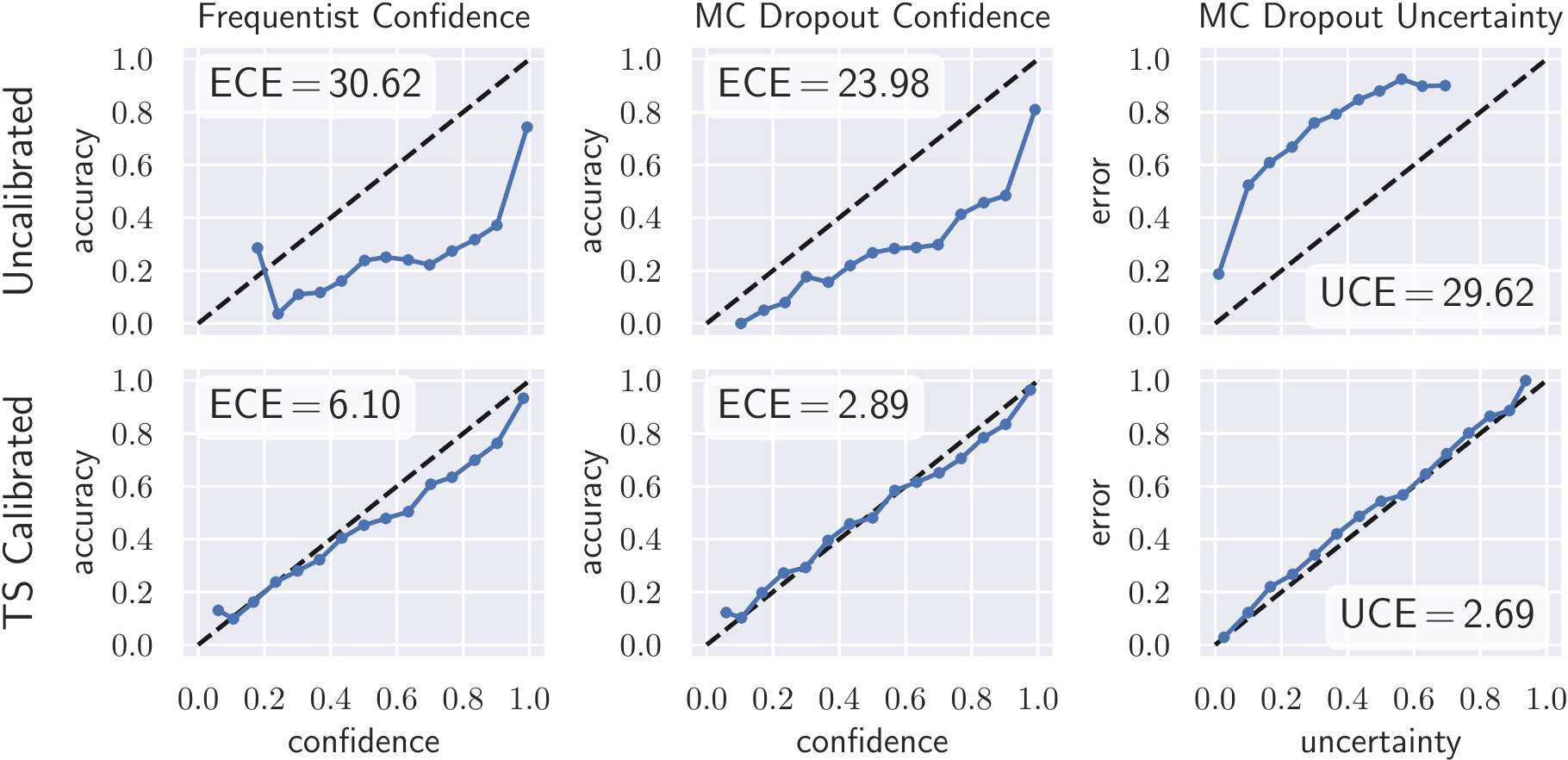}
    \caption{Reliability diagrams ($ M = 15 $ bins) for DenseNet-169 on CIFAR-100.}
    \label{fig:reliability_densenet}
\end{figure}

\end{document}